# A New Quantum CNN Model for Image Classification

Xingqiang Zhao[1], and Tianlong Chen[2]*

[1] School of Computer Science and Engineering, Sun Yat-Sen University, Guangzhou, China
xingqiangzhao0824@163.com

[2] Department of Science and Technology Service, Zhejiang Rural Commercial United Bank Co., LTD, Hangzhou, China, chentianlong150@163.com

**Abstract.** Quantum density matrix represents all the information of the entire quantum system, and novel models of meaning employing density matrices naturally model linguistic phenomena such as hyponymy and linguistic ambiguity, among others in quantum question answering tasks. Naturally, we argue that the quantum density matrix can enhance the image feature information and the relationship between the features for the classical image classification. Specifically, we (i) combine density matrices and CNN to design a new mechanism; (ii) apply the new mechanism to some representative classical image classification tasks. A series of experiments show that the application of quantum density matrix in image classification has the generalization and high efficiency on different datasets. The application of quantum density matrix both in classical question answering tasks and classical image classification tasks show more effective performance.

**Keyword:** Quantum model, Quantum density matrix, Image classification, CNN.

## 1 Introduction

In a quantum system, the quantum density matrix represents all the information of the entire quantum system. Earlier work has already demonstrated potential quantum advantage for natural language processing (NLP) [1] in a number of manners, and novel models of meaning employing density matrices naturally model linguistic phenomena such as hyponymy and linguistic ambiguity, among others [2].

Although some quantum methods have been verified practicable in NLP, it has not been used in computer vision (CV) [3]. Motivated by the quantum idea, we argue that the quantum density matrix can also enhance the image feature information and the relationship between the features for the classical image classification [4].

Specifically, we (i) combine density matrices and CNN to design a new mechanism that is called QIM for convenience; (ii) apply the new mechanism to some representative classical image classification tasks. Our experiments on Intel dataset [5] have verified our conjecture that it can also enhance the image feature information and the relationship between the features by applying the quantum density matrix method to the classical image classification tasks. Besides, a series of experiments on Mnist [6], Fashion-Mnist [7], Cifar 10/100 [8], ......, have shown that the quantum density matrix fits different datasets in image classification tasks.

## 2 A New Quantum Neural Network for Image Classification

### 2.1 Image Classification

Image classification, one of the main applications of CV, is the task of assigning labels to images through image analysis. The process of traditional image classification algorithms (refer to Figure 1 (a)) consists of two steps: (i) extracting features manually from the image; (ii) making use of the training support vector machine (SVM) [9] or other classifiers to execute image classification according to the extracted features. In general, the accuracy of the whole system of traditional image classification algorithms is limited to the method of extracting features. To solve the problem of feature extraction, AlexNet [10] is designed by applying deep learning [11] to large-scale image classification tasks and achieves better results. After that, many methods based on AlexNet emerged, such as LeNet [12], ZFNet [13], VGG-Net [14], ResNet [15], and so on.

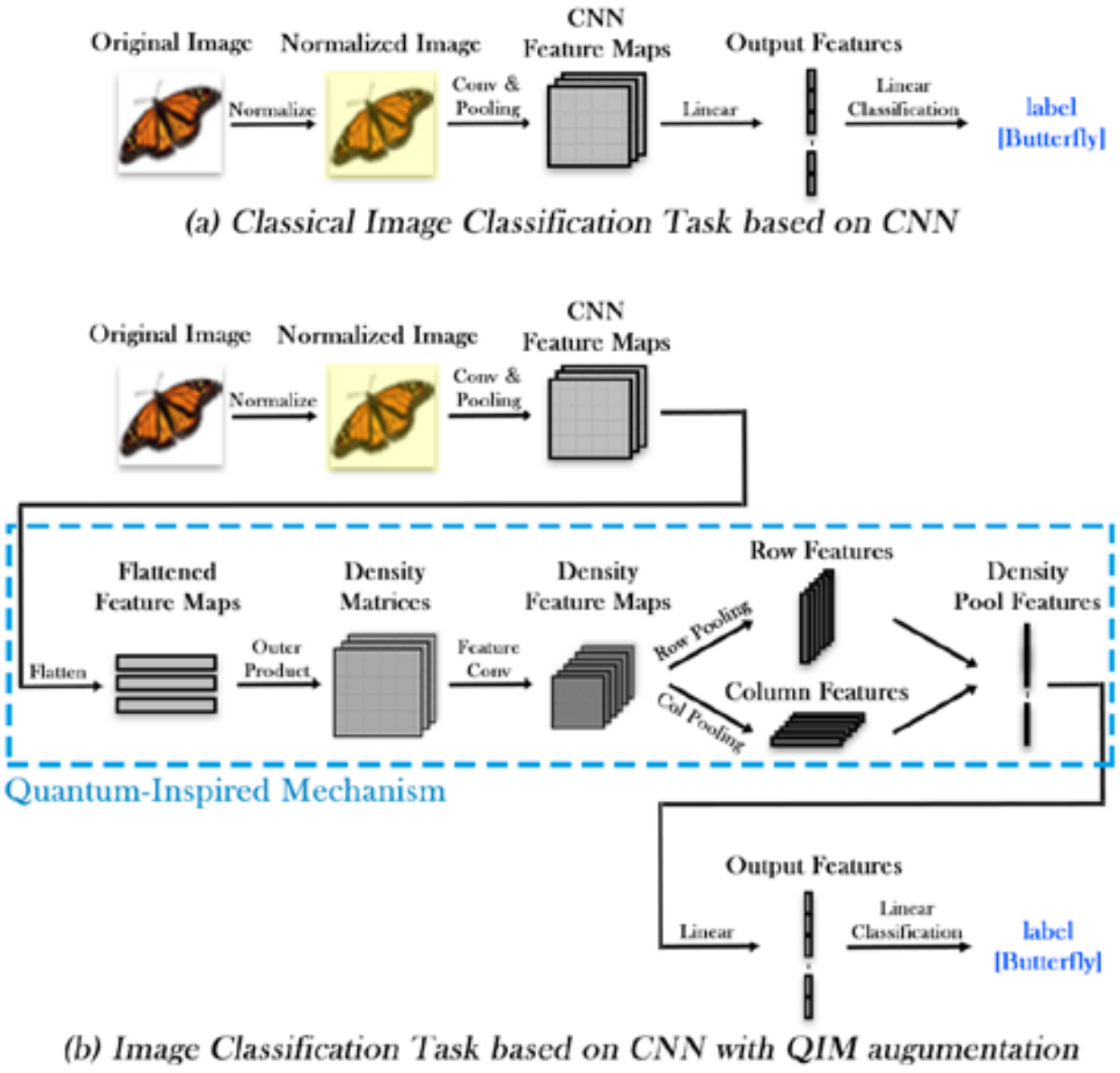

*(a) Classical Image Classification Task based on CNN*

*(b) Image Classification Task based on CNN with QIM augumentation*

**Fig. 1.** The process of classical image classification task based on CNN is shown in (a) and the process of image classification task based on CNN with QIM is shown in (b).

### 2.2 Scheme Design

We consider making the density matrix to enhance image features and the correlation information between features in classical image classification tasks. In a quantum system, every particle $i$ is in its own state $|s_i\rangle$, and the tensor product $|s_i\rangle\langle s_i|$ is its own quantum density matrix. The quantum density matrix of the entire system is generated by adding the quantum density matrix of each particle in the quantum system according to a certain weight $p_i$, the generation of quantum density matrix $\sum_i p_i |s_i\rangle\langle s_i|$ about the entire system is shown in Figure 2.

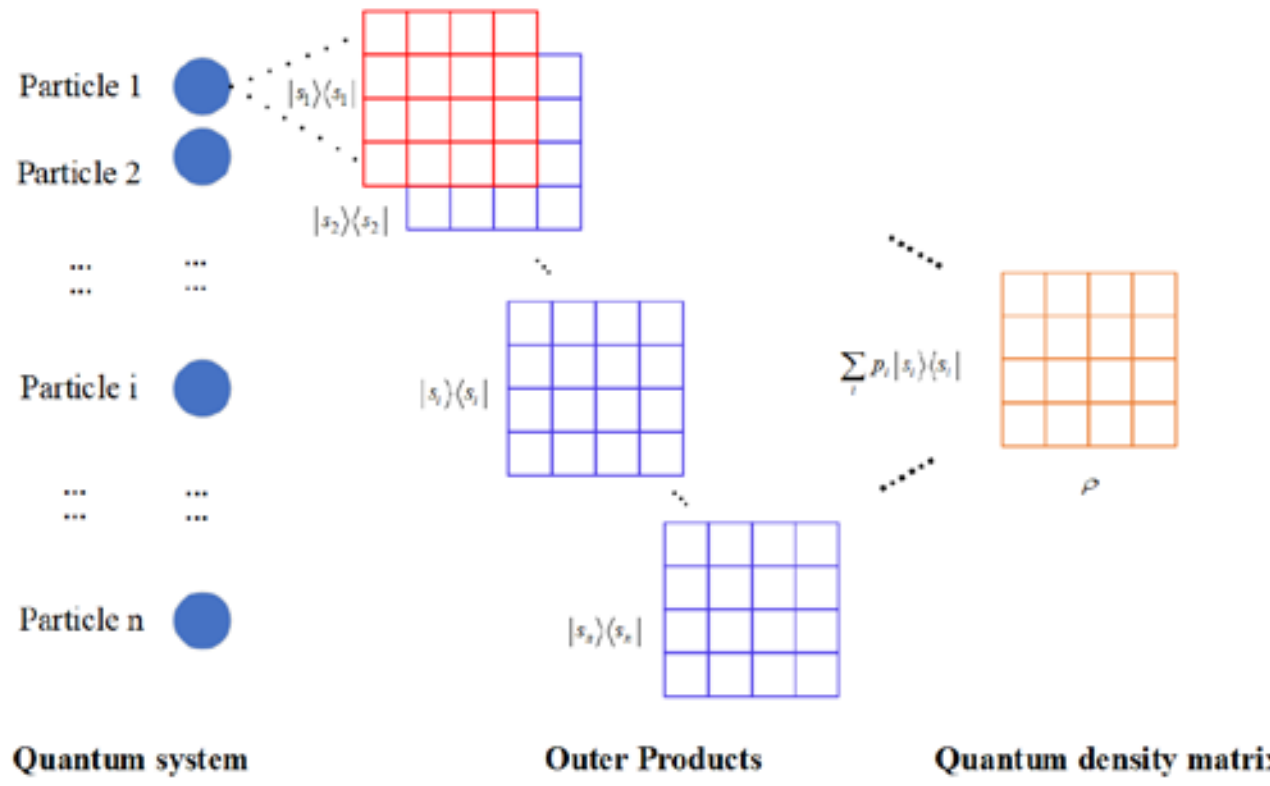

**Fig. 2.** Generation of quantum density matrix in quantum system.

Considering that the inputs of image information are pixel matrices in CV, we directly flatten these extracted feature matrices to obtain their own flattened feature map vectors $\{\vec{m}_i\}$. Then we perform the operation shown in Equation (1) on each $\vec{m}_i$ to obtain the corresponding density matrix $p_m^i \vec{m}_i \cdot \vec{m}_i^{\,T}$ with its own weight $p_m^i$. The generation process of density matrices is shown in Figure 3.

$$\rho^{'} = \sum_i k_i \vec{u} \cdot \vec{u}^{T} \tag{1}$$

Flatten

CNN feature map

Flattened feature map

Density matrices

**Fig. 3.** Generation of density matrices.

Finally, we add density matrices to CNN with 2D convolution kernels to generate our scheme of employing quantum density matrix in classical image classification tasks.

We name our scheme as Quantum Inspired Mechanism (QIM) for convenience, and the QIM is shown in Figure 4.

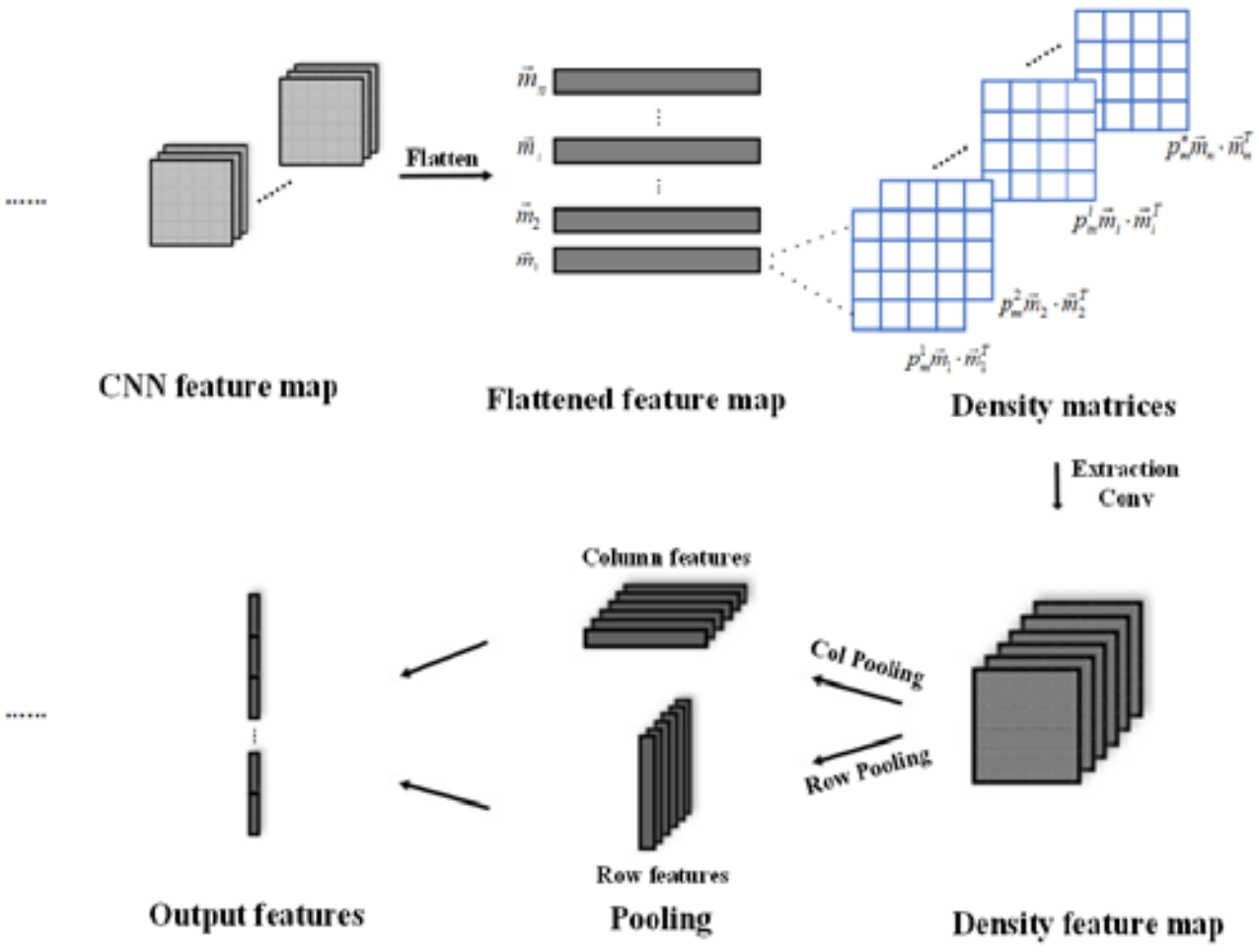

**Fig. 4.** The quantum inspired mechanism (QIM)

Suppose the number of filters is $c$. The $j^{th}$ convolution operation is

$$C_j = \delta\left[\left(\vec{m}_j \cdot \vec{m}_j^{\,T}\right) * p_m^j + b_j\right] \tag{2}$$

where $1 \le j \le c$, $\delta$ is the non-liner activation function, $*$ is the 2D convolution, $p_m^j$ and $b_j$ are the weight and bias respectively for the $j^{th}$ convolution kernel, and $C_j$ is the feature map. After the convolution layer obtains the feature maps, we then use row-wise and column-wise max-pooling to generate vectors $\vec{ro}_t \in R^{d-k+1}$ and $\vec{co}_t \in R^{d-k+1}$, respectively.

$$\vec{ro}_t = \max\left(C_{j,g}\right), 1 \le g \le k-d+1,\ \vec{co}_t = \max\left(C_{g,j}\right), 1 \le g \le k-d+1 \tag{3}$$

We concatenate these vectors as follow:

$$\vec{f} = \left[\vec{co}_1;\vec{ro}_1;\vec{co}_2;\vec{ro}_2;\ldots;\vec{co}_t;\vec{ro}_t;\ldots;\vec{co}_c;\vec{ro}_c\right] \tag{4}$$

where $1 \le t \le c$. The above convolution operation aims to extract useful similarity patterns and each convolution kernel corresponds to a feature. We can adjust the parameters of the kernels when the model is being trained.

### 2.3 Experimental Evaluation

To verify our scheme, we apply QIM to the classical image classification tasks. The specific process of the classical image classification task based on CNN with quantum density matrix is shown in Figure 1 (b), where QIM is marked blue. We design a series of preliminary ablation experiments on Intel dataset to test whether the density matrices enhance the image feature information and the relationship between the features. Furthermore, we design a series of ablation experiments according to the current better results achieved by typical models on different datasets to verify whether QIM can improve the accuracy of various image classification tasks.

We choose the image classification datasets by considering the following reasons: they have covered both gray-scale and colored images; both small-sized (28x28, 32x32) images and large-sized (150x150) images are considered; categories of image data range from 6 to 100.

To verify whether QIM is helpful for kinds of classical image classification tasks with some models on different datasets, we only add QIM to the current model with the better results, and the comparison results between these better results and the results obtained by adding QIM to these typical models are shown in Table 1. As for Cifar series, Cifar10 dataset contains small-sized images, as well as smaller classification space compared to Cifar 100, so we do not apply large models (e.g. VGG, ResNet) on it. StandardCNN [16] is a CNN model carried out for our baseline experiments. Its structure is designed in particular, including 3 convolutional layers, 3 pooling layers and 2 fully connected layers.

**Table 1.** Comparison results between typical models with the models adding QIM.

| Datasets | Image Size | Category Number | Approach | Batch Size | Learning Rate | Accuracy | Comparison |
|---|---|---|---|---|---|---|---|
| Mnist | 28x28 | 10 | StandardCNN | 64 | 0.0005 | 97.2622 | +1.8496 |
| | | | StandardCNN+ QIM | 64 | 0.0005 | 99.1118 | |
| | | | LeNet | 64 | 0.0005 | 99.1209 | +0.0373 |
| | | | LeNet+QIM | 64 | 0.0005 | 99.1582 | |
| | | | AlexNet | 64 | 0.0005 | 99.3395 | +0.1813 |
| | | | AlexNet+QIM | 64 | 0.0005 | 99.4943 | |
| | | | ZFNet | 64 | 0.0005 | 99.3531 | +0.2005 |
| | | | ZFNet+QIM | 64 | 0.0005 | 99.5536 | |
| Fashion-Mnist | 28x28 | 10 | LeNet | 64 | 0.0005 | 91.3962 | +0.1425 |
| | | | LeNet+QIM | 64 | 0.0005 | 91.5117 | |
| | | | AlexNet | 64 | 0.0005 | 92.0373 | +0.4006 |
| | | | AlexNet+QIM | 64 | 0.0005 | 92.4379 | |
| | | | ZFNet | 64 | 0.0005 | 91.9371 | +0.4107 |
| | | | ZFNet+QIM | 64 | 0.0005 | 92.3478 | |
| Cifar 10 | 32x32 | 10 | StandardCNN | 64 | 0.0005 | 67.5381 | +8.3834 |
| | | | StandardCNN+QIM | 64 | 0.0005 | 75.9215 | |
| | | | AlexNet | 64 | 0.0005 | 78.5757 | +1.1919 |
| | | | AlexNet+QIM | 64 | 0.0005 | 79.7676 | |
| | | | ZFNet | 64 | 0.0005 | 80.1983 | +0.3452 |
| | | | ZFNet+QIM | 64 | 0.0005 | 80.5435 | |
| Cifar 100 | 32x32 | 100 | AlexNet | 64 | 0.0005 | 64.5532 | +0.4708 |
| | | | AlexNet+QIM | 64 | 0.0005 | 65.0240 | |
| Intel | 150x150 | 6 | StandardCNN | 64 | 0.0005 | 79.4387 | +3.5097 |
| | | | StandardCNN+QIM | 64 | 0.0005 | 82.9484 | |
| | | | AlexNet | 64 | 0.0005 | 77.6835 | +7.0651 |
| | | | AlexNet+QIM | 64 | 0.0005 | 84.7486 | |
| | | | ZFNet | 64 | 0.0005 | 83.6956 | +1.8000 |
| | | | ZFNet+QIM | 64 | 0.0005 | 85.4959 | |
| | | | VGG | 16 | 0.0001 | 76.2433 | +6.6196 |
| | | | VGG+QIM | 16 | 0.0001 | 82.8629 | |
| | | | ResNet | 32 | 0.00025 | 80.8763 | -0.6304 |
| | | | ResNet+QIM | 32 | 0.00025 | 80.2459 | |

It is obviously that most models with QIM have higher accuracy on different datasets than before. QIM can significantly improve the accuracy of original models (StandardCNN, LeNet, AlexNet, ZFNet, VGG, and ResNet) with better results on different datasets (Mnist, Fashion-Mnist, Cifar 10, Cifar 100, and Intel). All the experiments show that QIM is really helpful for classical image classification tasks and it is applicable with above models on different datasets. The performance of model without/with QIM on different datasets is shown in Figure 5, where the abscissa represents different image datasets, and the ordinate represents image classification accuracy with different models. It is clear that QIM does help to improve the performance in majority of cases.

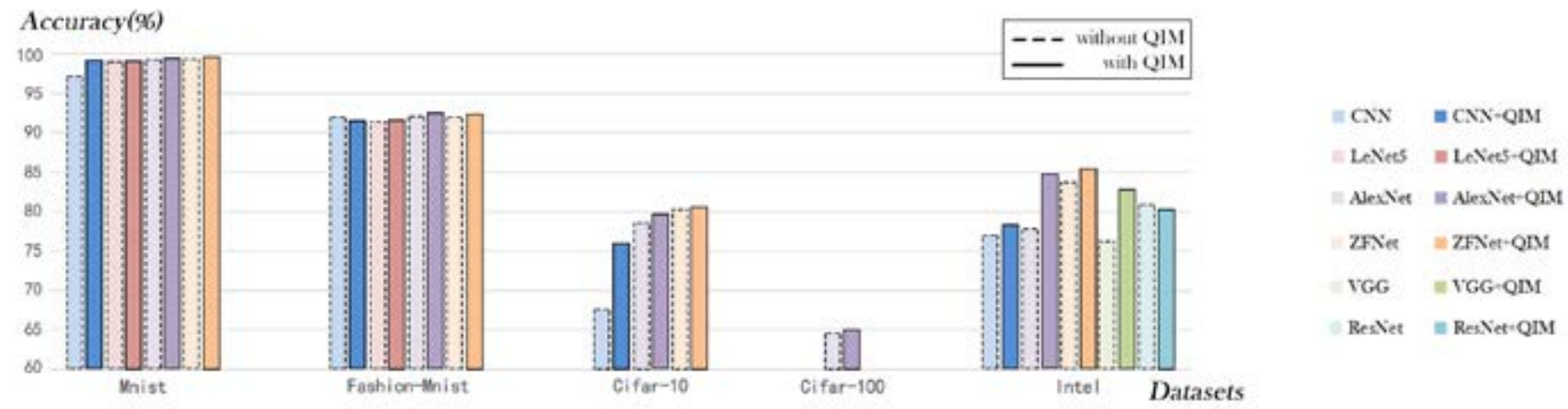

**Fig. 5.** Accuracy of models without/with QIM on different datasets of classifiers trained.

In most cases, image classifiers gained higher accuracy when attached to QIM. Besides, we have noticed that QIM benefits more when original CNN has difficulty in extracting image features effectively. However, there is still another exception. For example, applying ResNet-18 to Intel dataset, QIM has made it more difficult in achieving higher accuracy. A possible reason is that QIM emphasizes mainly on carrying CNN models to extract more detailed information, without making original CNN models much deeper or wider. So QIM may not work well on those tasks in which the original model itself fits well in extracting image information. We stress that QIM is a proof-of-concept, and there are many possible optimizations but they are beyond the scope of this paper.

## 3 Conclusion

To investigate whether the quantum density matrix helps the classical image classification tasks, we have applied the quantum density matrix into classical image classification tasks and proposed the quantum inspired mechanism, i.e., QIM, which re-represented the quantum density matrix with linear algebra and combined CNN with density matrices. To the best of our knowledge, this is the first time for the quantum method applied to image classification tasks. In QIM, density matrices enhance image features and the correlation between features in classical CV algorithms, which has been verified by a series of preliminary ablation experiments.

We gained the optimum parameters for QIM from preliminary ablation experiments and applied these parameters to the further ablation experiments. Results have shown that QIM has the generalization in classical image classification tasks on different datasets. To sum up, the application of quantum density matrix in classical image classification tasks have shown more effective performance, which confirms our argue about the quantum density matrix.